\documentclass[sigconf]{aamas} 

\usepackage{balance} 

\usepackage{multirow,array}
\usepackage{subfig}
\usepackage{graphicx}
\usepackage{amsmath} 

\usepackage{comment}

\DeclareMathOperator*{\argmax}{arg\,max}

\usepackage{booktabs}

\setcopyright{ifaamas}
\acmConference[AAMAS '22]{Proc.\@ of the 21st International Conference
on Autonomous Agents and Multiagent Systems (AAMAS 2022)}{May 9--13, 2022}
{Auckland, New Zealand}{P.~Faliszewski, V.~Mascardi, C.~Pelachaud,
M.E.~Taylor (eds.)}
\copyrightyear{2022}
\acmYear{2022}
\acmDOI{}
\acmPrice{}
\acmISBN{}

\acmSubmissionID{504}

\title{Active Altruism Learning and Information Sufficiency for Autonomous Driving}

\author{Jack Geary}
\affiliation{
  \institution{School of Informatics, University of Edinburgh,}
  \city{Edinburgh}
  \country{UK}}
\email{jack.geary@ed.ac.uk}

\author{Henry Gouk}
\affiliation{
  \institution{School of Informatics, University of Edinburgh,}
  \city{Edinburgh}
  \country{UK}}
\email{herny.gouk@ed.ac.uk}

\author{Subramanian Ramamoorthy}
\affiliation{
  \institution{University of Edinburgh,}
  \city{Edinburgh}
  \country{UK}}
\affiliation{
  \institution{FiveAI Ltd.,}
  \country{UK}}
\email{s.ramamoorthy@ed.ac.uk}

\begin{abstract} 
Safe interaction between vehicles requires the ability to choose actions that reveal the preferences of the other vehicles. Since exploratory actions often do not directly contribute to their objective, an interactive vehicle must also able to identify when it is appropriate to perform them. In this work we demonstrate how Active Learning methods can be used to incentivise an autonomous vehicle (AV) to choose actions that reveal information about the altruistic inclinations of another vehicle. We identify a property, Information Sufficiency, that a reward function should have in order to keep exploration from unnecessarily interfering with the pursuit of an objective. We empirically demonstrate that reward functions that do not have Information Sufficiency are prone to inadequate exploration, which can result in sub-optimal behaviour. We propose a reward definition that has Information Sufficiency, and show that it facilitates an AV choosing exploratory actions to estimate altruistic tendency, whilst also compensating for the possibility of conflicting beliefs between vehicles.
\end{abstract}


\begin{document}


\pagestyle{fancy}
\fancyhead{}


\maketitle 


\section{Introduction}
\begin{figure}[!h]
    \centering
    \subfloat[\label{intro_figure_graphic}]{%
        \includegraphics[height=.22\textheight]{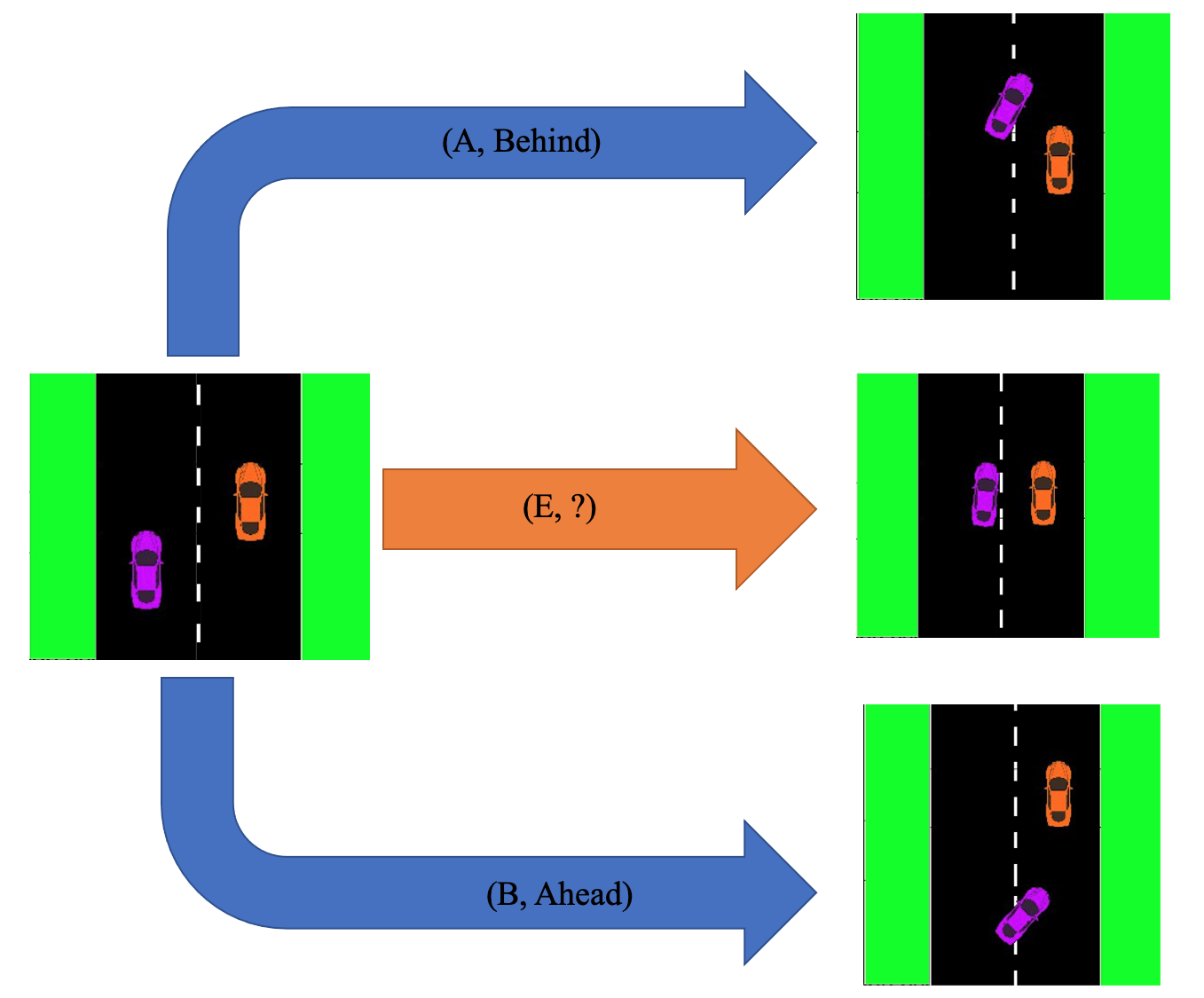}}
    \begin{center}
        \subfloat[\label{intro_figure_table}]{%
            \begin{tabular}{cc|c|c|}
                & \multicolumn{1}{c}{} & \multicolumn{2}{c}{$C$}\\
                & \multicolumn{1}{c}{} & \multicolumn{1}{c}{Behind}  & \multicolumn{1}{c}{Ahead} \\\cline{3-4}
                \multirow{2}*{$R$}  & $A$ & $(1,0)$ & $(-1,-1)$ \\\cline{3-4}
                & $B$ & $(-1,-1)$ & $(0,1)$ \\\cline{3-4}
                & $E$ & $(1,0)$ & $(0,1)$ \\\cline{3-4}
            \end{tabular}}
    \end{center}
\caption{Lane change decision-making scenario; (a) The Row player (R, Purple) wants to merge into the Column player's (C, Orange) lane. Both players would prefer to be ahead of the other after the manoeuvre is completed. R can attempt to merge ahead ($A$), merge behind ($B$), or perform an exploratory action ($E$). C can choose to give way (Behind), or they can stay ahead of R (Ahead). (b) Reward matrix associated with the scenario.}
\label{intro_figure}
\end{figure}

Interaction arises in driving scenarios where vehicles must coordinate their behaviours in order to complete their objectives. Scenarios involving crossing an unsignalised intersection or merging into an occupied lane can require interaction between vehicles in order to be safely executed. In these scenarios the optimal behaviour of each vehicle can depend on the behaviours the other vehicles intend to execute. Therefore it is necessary that an autonomous vehicle (AV) in this setting be able to accurately estimate the preferences of the other vehicles and anticipate their intentions.

Preferences and intents are typically communicated through behaviour; drivers have access to explicit communication methods, such as turn signals, to communicate their intent. However, it is not uncommon for communication to be entirely implicit, with drivers conveying their preferences via their actions. For example, a driver may choose to accelerate in order to stop another vehicle from merging ahead of them. The ability to reliably interpret implicit signals can be integral to the execution of certain driving manoeuvres; highway drivers have been shown to be able to estimate when a vehicle intends to perform a lane change, without relying on an explicit communication signal \cite{Driggs-Campbell2016}. This is an ability that is lacking in current autonomous driving systems. Therefore, for AVs to be able to interact effectively with human-controlled vehicles they must be able to infer the preferences and intents of other drivers without relying on explicit communication channels.

When decision-making is motivated by inference, an AV's decisions can be motivated by the accuracy of the inferred value, for example choosing conservative actions in order to gather more observations. This can result in the AV behaving too defensively, or failing to infer anything at all. In order to address this, an AV needs to be able to identify when they have sufficient information to make a decision, and when further inference is necessary, and be able to choose actions that will provide evidence to support the inference if required.

In \cite{geary2020,geary2021} the authors model interactive driving scenarios as Stackelberg Games between the planning vehicle and the other vehicles on the road \cite{von1934}. Each player has an associated parameter, $\alpha$, that indicates the player's interest in the other players achieving their objectives. This formulation relates to the Game Theoretic notion of altruism \cite{andreoni1993}, as well as Social Value Orientation (SVO)-based models proposed in \cite{Schwarting2019} which have garnered recent interest \cite{toghi2021a,toghi2021b}. They show how this formulation can be used to efficiently compute interactive policies that account for each player's preferences. However, in their experiments the value of $\alpha$ for each player was presumed to be known a priori. This is an unrealistic requirement in the open world, and any AV that might use such a model would need to be able to infer this parameter value. Previous works have explored methods for performing inference on similar parameters (e.g., \cite{Schwarting2019}). However, these approaches are often ``passive'', in that they wait and see what a vehicle does, and uses this observational data to inform the inference. In practice, since these parameters affect how agents interact, often their value can only be accurately inferred through interaction. This requires the planning vehicle to choose an action and observe how the other vehicles respond. Relying on ``passive'' inference approaches may fail to infer the value, since the data gathered may lack any interaction and, therefore, be incomplete. These approaches can also result in overly passive behaviour by the AV.

In this work we demonstrate how Active Learning and Information Gathering methods can be incorporated into the model proposed in \cite{geary2021} in order to infer the value of $\alpha$. Using these approaches the planning vehicle's reward function is augmented to incentivise actions that would reveal information about $\alpha$, even if they do not directly pursue the planning vehicle's objective. We also demonstrate that our proposed method requires fewer assumptions on the reward matrix values than other methods, and that it can even operate when the players have conflicting beliefs over the role of leader and follower \cite{geary2021}. 

Typical Active Learning methods can result in an AV choosing actions in order to gain information that does not contribute to their objective. As such, it is important that any method that utilises Active Learning approaches is also able to determine when there is nothing more to be gained from further exploration. To this end we define \emph{Information Sufficiency}, a property that a reward function should have in order for it to be suitable for use in Active Learning. We demonstrate that Active Learning approaches motivated by information gain do not have information sufficiency, and show that this can result in inadequate exploration and sub-optimal behaviour. We propose a novel alternative to information gain that does have the information sufficiency property, and we compare the performance of the two models.

The key contributions of this work are as follows:
\begin{itemize}
    \item Incorporating Active Learning methods in order to infer the value of $\alpha$ in an Stackelberg Game-based model for interactive decision-making.
    \item Identifying Information Sufficiency, a property a reward function should have in order to avoid unnecessary exploration. We also propose a novel reward function that has this property and demonstrate its effectiveness in a lane merge scenario.
    \item Demonstrating how Conflict-awareness \cite{geary2021} can be incorporated into the decision-making, thereby reducing the assumptions required on the reward function.
\end{itemize}

\section{Active Altruism Learning for Stackelberg Games}
In this section we will introduce the Stackelberg Game formulation as well as the incorporation of Altruism as proposed in \cite{geary2021}. We subsequently detail how Active Learning methods can be used to infer an unknown value for the altruism coefficient, $\alpha$, in this setting. 
\subsection{Altruism for Stackelberg Games}
A Stackelberg (Leader-Follower) Game, \cite{von1934}, is a simultaneous game formulation between two players where one player assumes the role of Leader, and the other the Follower. The Leader chooses the action that results in the best outcome for them, and the Follower chooses their best action, conditioned on the Leader's expected action choice. The equilibrium for such games is unique and can be efficiently computed, making it very suitable for applications such as autonomous driving. 

In \cite{geary2021} the authors address the autonomous driving planning problem with a hierarchical model; a decision-making model is used to determine what type of trajectory to execute, and a planning model is used to determine how it should be executed. Decision-making is formulated as a Bimatrix Stackelberg Game (e.g. Figure \ref{intro_figure_table}). In this context each player has a discrete set of actions which are known, and each action combination $(Ai,Bj)$ has associated with it a pair of rewards $(r_{ijR},r_{ijC})$ which would be received by the row and column players, respectively, if that combination of actions were realised. It is assumed that the row player, $R$, is the leader, and that the column player, $C$, will break ties in favour of the leader. 

The altruistic tendency of a player, $a$, is identified by a coefficient, $\alpha_{a} \in [0,1], a \in \{R,C\}$ which reweights the reward received by the player according to:
\begin{equation}
    r_{a}(Ai,Bj,\alpha) = (1-\alpha_{a})r_{ija} + \alpha_{a}r_{ij(-a)}
\end{equation}
where $-a$ indexes the player that is not $a$. Values closer to $0$ indicate that the player is more motivated by achieving their own reward and is incentivised to behave selfishly, whereas values closer to $1$ indicate the player is incentivised by the other player's success, and are motivated to behave more altruistically. In this work we assume that the leader is the planning (ego) agent, and the value $\alpha_{R}$ is known. For simplicity, we let $\alpha_{R}=0$, and $\alpha=\alpha_{C}$. 
\subsection{Parameter Inference for Stackelberg Games}
\label{parameter_inference_for_stackelberg_games}
In \cite{geary2021} the authors presumed the value of $\alpha$ was known. In practice this is unlikely to be the case. Instead, the leader would maintain a belief over $\alpha$, $b(\alpha)$, and their decision-making objective would be to choose the action that maximises the expected reward,
\begin{equation}
    A^{*} = \argmax_{\tilde{A} \in A} \int_{\alpha} b(\alpha)R_{R}(\tilde{A},\alpha),
\end{equation}
where $R_{R}(A,d)$ returns the reward the $R$ would receive for choosing action $A$ when $\alpha=d$, under the assumption that $C$ (the follower in the Stackelberg game) will respond rationally. In this setting, if $R$ observes the equilibrium $(Ai,Bj)$ they can conclude that:
\begin{equation}
    \begin{split}
        r_{C}&(Ai,Bj,\alpha) \geq r_{C}(Ai,Bk,\alpha) \forall Bk \in B \\
        \implies & \frac{r_{ijC}-r_{ikC}}{r_{ikR}-r_{ikC}+r_{ijC}-r_{ijR}} \geq \alpha \forall k \neq j
    \end{split}
\end{equation}
where $B$ is the set of actions available to $C$, and $r_{C}$ returns the reward to $C$ for the specified action combination and value of $\alpha$. More generally, every pair of possible action combinations $(Ai,Bj),(Ai,Bk)$ divides the domain of $\alpha$ about the point of intersection of the corresponding reward functions $r_{C}(Ai,Bj,\alpha),r_{C}(Ai,Bk,\alpha)$. 

Initially $R$ has no information about the value of $\alpha$, so $b^{0}(\alpha) = U(0,1)$, the uniform distribution over the range $\alpha \in [0,1]$. Each iteration, $t$, of decision-making defines a range in the domain, $[c,d] \subseteq [0,1]$, in which the ``true" value for $\alpha$ lies. By determining the intersection of this range with the existing belief over the range of possible values, $[c^{t},d^{t}]$, we can generate an updated range within which the ``true" value exists. We can then use this information to update the belief over $\alpha$:

\begin{equation}
    \begin{split}
        c^{t+1} &= \text{max}(c^{t},\text{min}(d^{t},c))\\
        d^{t+1} & = \text{min}(d^{t},\text{max}(c^{t},d))\\
        b^{t+1}(\alpha) &= \begin{cases}
        U(c^{t+1},d^{t+1}) &\alpha \in [c^{t+1},d^{t+1}] \\
         0 &\alpha \notin [c^{t+1},d^{t+1}]
        \end{cases}
    \end{split}
\end{equation}

We refer to this type for parameter inference as \emph{Passive}, since the conclusions drawn about $\alpha$ are incidental to the decision-making that would occur even if no inference were occurring. While this method can converge on reasonable ranges bounding $\alpha$, it is also possible for this method to fail entirely; for instance, if at iteration $t$ the leader chooses an uninformative action, such that $[c^{t},d^{t}] \subseteq [c,d]$, then the bounds do not change. As a consequence, the same action will be chosen in every subsequent round of decision-making, so no further progress will be made in inferring $\alpha$. In order to motivate progress on the inference task, \emph{Active} methods should be used.

\subsection{Active Information Gathering in Stackelberg Games}
In Active Learning our goal is to incorporate parameter inference into action selection to improve the inference quality. One such approach is to identify a function, $J$, to augment, $R_{R}$, in order to motivate choosing actions that might provide more information about the parameter of interest,$\alpha$; $R_{R}^{*} = R_{R} + J$ \cite{sadigh2016}. Compared to $R_{R}$, $R_{R}^{*}$ motivates choosing more informative actions, which reduces the possibility of scenarios where the agent learns nothing about $\alpha$.

In \cite{sadigh2018,Furlanis2019} the authors use information gain to augment $R_{R}$; $J = \lambda (H(b^{t})-H(b^{t+1}))$ where $H$ is the entropy:
\begin{equation}
    H(b) = -\frac{\sum_{\alpha} b(\alpha)\log(b(\alpha))}{\sum_{\alpha}b(\alpha)}
\end{equation}
 
$\lambda$ determines the degree to which $R$ is motivated by the information gathering task; if $\lambda$ is low, then the agent will not be incentivised to choose informative actions. But if $\lambda$ is too high, then $R$ could disregard their objective, and choose actions that are maximally informative, but also suboptimal or dangerous (\cite{sadigh2018} discuss this in greater detail). Determining how best to address this trade-off between exploring informative actions, and exploiting rewarding actions, is a topic of recent and ongoing research \cite{berger2014,Brooks2019}.

In \cite{sadigh2018,Furlanis2019}, these methods are applied at the planning level of the trajectory generation process. In this work we use a hierarchical trajectory generation process, and incorporate the active learning at the decision-making level instead. Unlike in the planning-level application, in our work we presume that the rewards associated with each action are fixed and known. As a result we can use an alternative approach for information gathering.

In practice, it is not necessarily best to gather as much information as possible, as is indicated by the entropy-based motivation. Instead the goal is to gather as much \emph{useful} information as possible; after a certain point learning ceases to be beneficial, since the resulting information does not further the planning agent's primary objective. This is not tied to the \emph{amount} of information that is unknown, but how \emph{consequential} the unknown information might be. With this motivation in mind, we propose an alternative definition for $J$; \emph{Expected Reward Gain}:

\begin{equation}
    \begin{split}
        F(b) &= \sum_{a \in A} \mathbb{E}_{\alpha \sim b}[R_{R}(a,\alpha)] \\
        J &= \lambda(|F(b^{t+1})-F(b^{t})|)
    \end{split}
\end{equation}

$F$ measures the expected reward to $R$ for a given distribution over the value of $\alpha$, so this definition of $J$ measures how much $R$ stands to gain from exploring as opposed to directly pursuing their objective. This is related to \emph{Advantage} in Reinforcement Learning, \cite{baird1994}, where the difference between the estimated expected reward from a given state $s$, $V(s)$ and the expected reward  from a particular action $a$, $Q(s,a)$, motivates the action selection. As opposed to information gain, where the scale of $J$ is independent of the magnitude of the $R_{R}$, the scale of the expected reward gain is scaled by the magnitudes of the rewards. As a result we would expect a reduced dependence on the magnitude of $\lambda$. Expected reward gain would also be expected to have a low magnitude if there is little to be gained from further exploration, even if there is still high uncertainty over $b$.

\subsection{Active Altruism Learning for Stackelberg Games}
\label{active_altruism_learning_for_stackelberg_games}
\begin{figure}[h]
    \begin{center}
        \subfloat[\label{info_gathering_example_table}]{%
            \begin{tabular}{cc|c|c|}
                & \multicolumn{1}{c}{} & \multicolumn{2}{c}{$C$}\\
                & \multicolumn{1}{c}{} & \multicolumn{1}{c}{$B1$}  & \multicolumn{1}{c}{$B2$} \\\cline{3-4}
                \multirow{2}*{$R$}  & $A1$ & $(3,0)$ & $(-5,7)$ \\\cline{3-4}
                & $A2$ & $(-1,2)$ & $(1,1)$ \\\cline{3-4}
                & $A3$ & $(-1,2)$ & $(2,2)$ \\\cline{3-4}
            \end{tabular}}
    \end{center}
    \caption{Reward Matrix for the Information Gathering example}
    \label{fig:information_gathering_example_reward_matrix}
\end{figure}

To demonstrate the different behaviours that arise from different definitions of $J$ we examine a simple example (Figure \ref{fig:information_gathering_example_reward_matrix}); action $A1$ has the highest magnitude cost (-5), but also the highest reward (3) for the $R$. Action $A3$ is the ``safe" action since, for any value of $\alpha$, $R$ will get a reward of $1$. However, this action is also the least informative, as $C$'s action provides no information about their altruistic inclination. Action $A2$ is the exploratory (``nudge" \footnote{In this work we will sometimes refer to the exploratory action as a ``nudge". This arises from \cite{sadigh2018} where the authors observed that one manifestation of exploratory behaviour in lane merges was the planning vehicle would ``nudge" into the other lane before performing the manoeuvre as a way of testing the other vehicle's response.}) action, as it does not have the highest expected reward ($\mathbb{E}_{\alpha \sim U(0,1)}[A2] = \frac{1}{3}$ whereas $\mathbb{E}_{\alpha \sim U(0,1)}[A3] = 2$), but it is informative (the intersection point of the two rewards is at $\alpha=\frac{1}{3}$) without assuming excessive risk.

\begin{figure}[t]
    \centering
    \subfloat[\label{fig:active_learning_exp1_expected_reward}]{%
        \includegraphics[width=.17\textwidth]{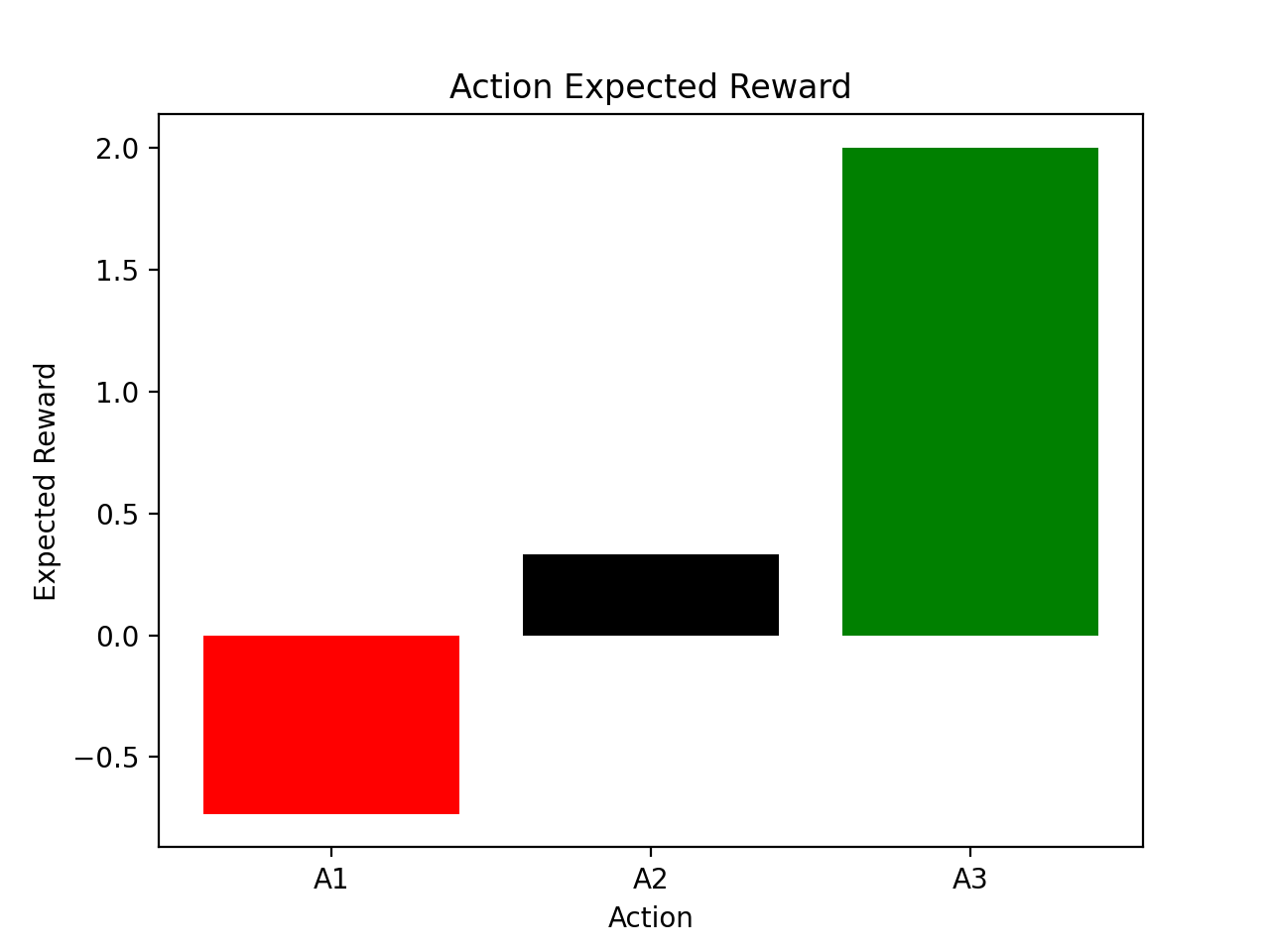}}
    \subfloat[\label{fig:active_learning_exp1_expected_info_gain}]{%
        \includegraphics[width=.17\textwidth]{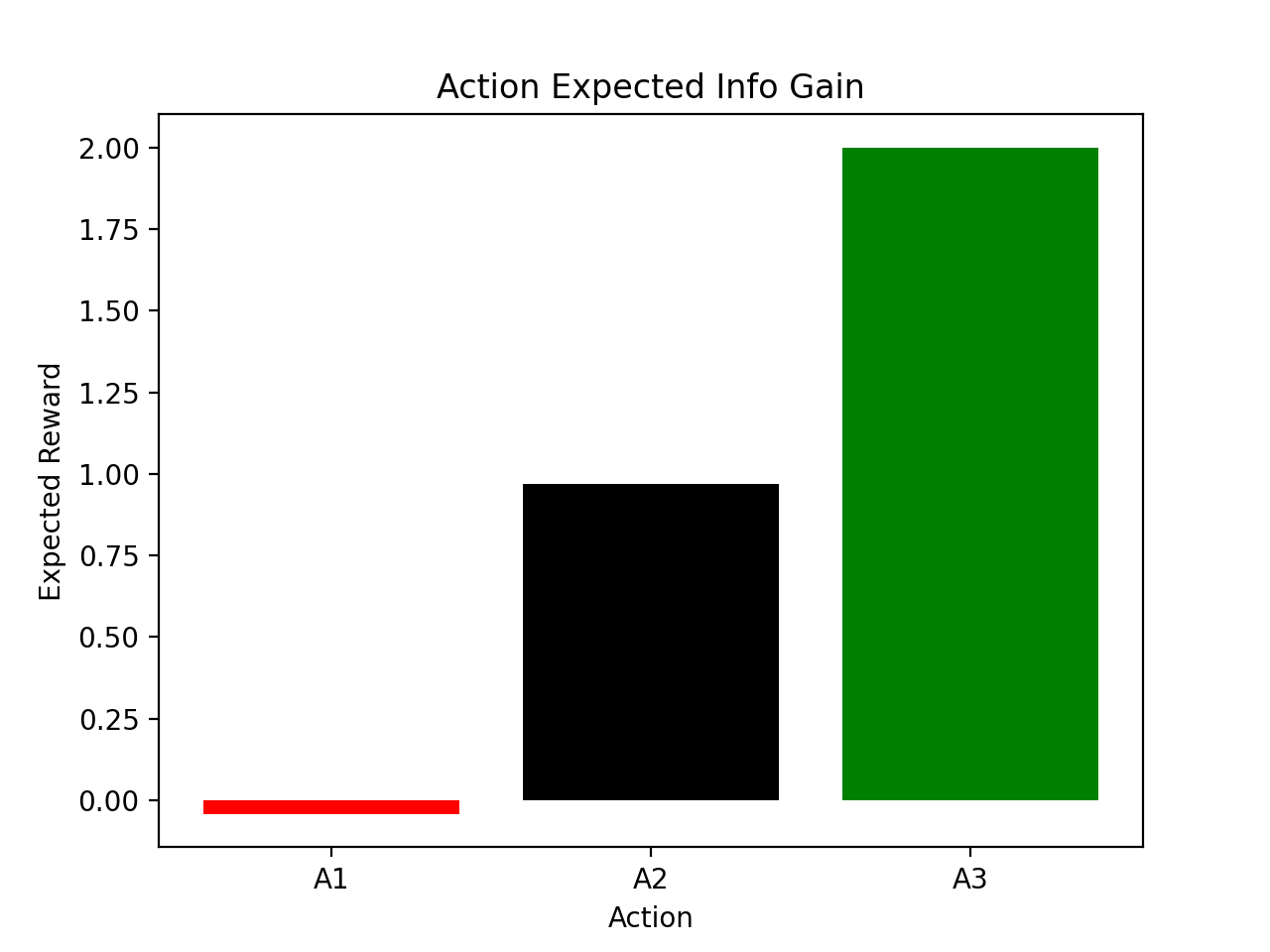}}
    \subfloat[\label{fig:active_learning_exp1_expected_reward_gain}]{%
        \includegraphics[width=.17\textwidth]{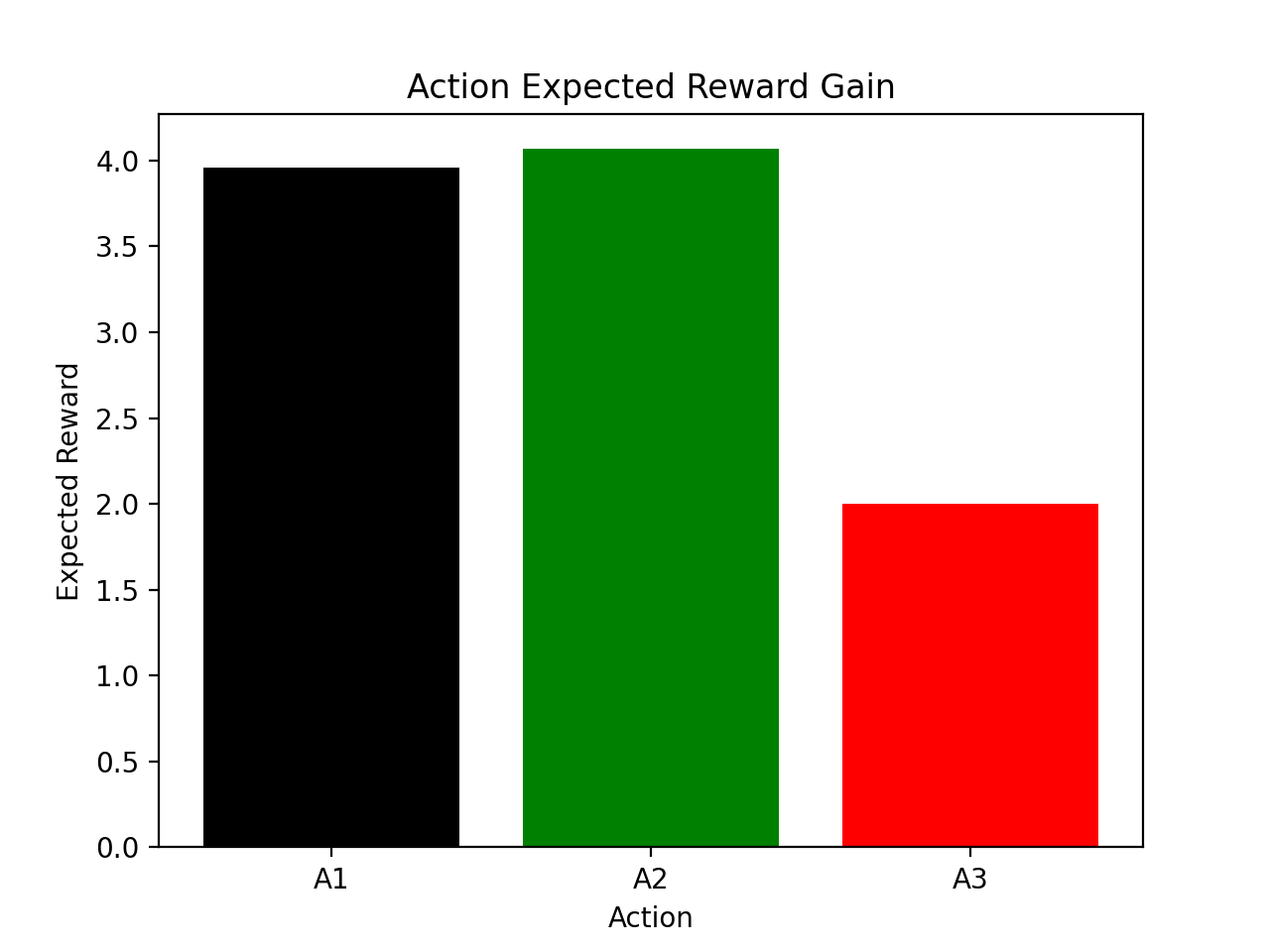}}
    \caption{Expected reward, based on the values in Figure \ref{fig:information_gathering_example_reward_matrix}, according to different definitions of $R^{*}_{R}=R_{R}+J$, computed from $b^{0}=U(0,1)$; (a) Passive update, $J=0$; (b) Information Gain, $J = H(b^{t})-H(b^{t+1})$; (c) Expected Reward Gain, $J = |F(b^{t+1})-F(b^{t})|$. The green and red bars identify the actions with the highest and lowest values respectively.}
\label{fig:active_learning_exp1_bar_plot}
\end{figure}

Figure \ref{fig:active_learning_exp1_bar_plot} depicts the expected values for $R_{R}^{*}$ for different definitions of $J$, starting from $b^{0} = U(0,1)$. We observe from Figure \ref{fig:active_learning_exp1_expected_reward} that, with no other information, $A3$ produces the highest immediate return. However, this action receives a reward of $2$ regardless of the value of $\alpha$, so this option provides no information. This is reflected in the fact that the magnitude of the $A3$ bar is fixed in each sub-figure in Figure \ref{fig:active_learning_exp1_bar_plot}. 

Figure \ref{fig:active_learning_exp1_expected_info_gain} depicts the expected returns when the reward is motivated by information gain; we observe that the values for both actions $A1$ and $A2$ have increased, but the value of $A3$ is unchanged. This is because both of those actions result in intersections within the domain of possible values of $\alpha$, which does not occur for $A3$. In spite of this, the magnitude of increase is not sufficient, and the ``safe" action is still preferred, so $R$ is not motivated to explore. 

Figure \ref{fig:active_learning_exp1_expected_reward_gain}, presents the expected returns when expected reward gain is used to augment the returns. In this case the augmented reward captures the tangible benefit associated to each action, not just the information learnt; even though $A1$ has a lot of associated risk, it also provides a lot of gain to the expected reward since, as a consequence, $R$ will either; i) get a reward of $3$ and know to consistently get this reward ($\alpha>\frac{7}{15}$) or ii) know that $A2$ is more likely to have $-1$ payoff ($\alpha<\frac{1}{3}$) than to have a payoff of $1$ ($\frac{1}{3}<\alpha<\frac{7}{15}$). Thus $A1$'s expected return now exceeds $A3$, for which there is no associated benefit. Similarly $A2$ received a spike in value, although not as significant as $A1$. But, as there was less risk associated with $A2$ initially, it comes out as being the preferred action. Therefore our method results in a ``nudge" behaviour selection, as observed in \cite{sadigh2018}. 

The results used in Figure \ref{fig:active_learning_exp1_bar_plot} were generated using $\lambda=1$. Alternative values could be used to demonstrate different results. This demonstrates the significance of the relationship between the effectiveness of this approach and the values of $R_{R}$ and the value of $J$; in the information gain definition $R_R$ is the expected value of rewards, which are scaled in relation to one another, and $J$ uses nats in its definition. In the expected reward gain definition, both $R_R$ and $J$ are defined using commonly scaled terms. The common units helps ensure that, $\lambda$, captures $R$'s exploration motivation, and does not have to compensate for differences in scale between units. We elaborate on the consequences of this on the overall behaviour in the next section. 
\section{Information Sufficiency}
\label{information_sufficiency}
\begin{figure}[h]
    \begin{center}
        \subfloat[\label{info_sufficiency_example_table}]{%
            \begin{tabular}{cc|c|c|}
                & \multicolumn{1}{c}{} & \multicolumn{2}{c}{$C$}\\
                & \multicolumn{1}{c}{} & \multicolumn{1}{c}{$B1$}  & \multicolumn{1}{c}{$B2$} \\\cline{3-4}
                \multirow{2}*{$R$}  & $A1$ & $(5,-4)$ & $(-2,1)$ \\\cline{3-4}
                & $A2$ & $(1,-4)$ & $(0,1)$ \\\cline{3-4}
            \end{tabular}}
    \end{center}
    \caption{Reward Matrix for the Information Sufficiency example}
    \label{fig:information_sufficiency_example_reward_matrix}
\end{figure}

\begin{figure}[t]
    \centering
    \subfloat[\label{fig:exp1_results}]{%
        \includegraphics[height=.2\textheight, width=.75\columnwidth]{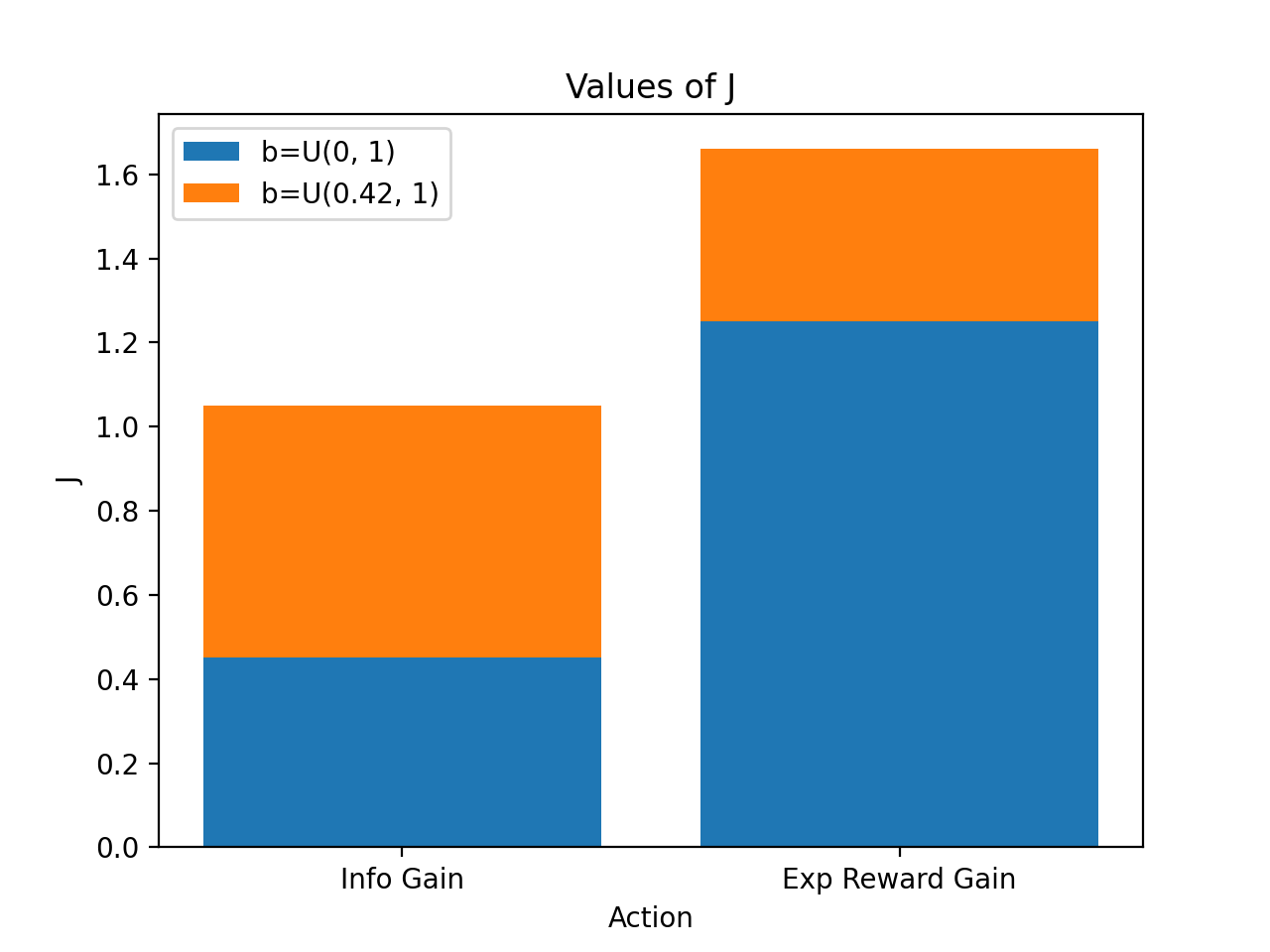}}
    \newline
        \subfloat[\label{table:exp1_results}]{%
        \begin{tabular}{llc}
            \toprule
            $b$                              & Information Gain   & Expected Reward Gain\\
            \midrule
            $U([0,1])$            & $[0.68,0.45]$               & $[3.54,1.25]$ \\
            $U([\frac{5}{12},1])$  & $[0,0.6]$               & $[0,0.41]$ \\
            \bottomrule
        \end{tabular}}
    \caption{Results from Experiment 1. (a) Stacked bar chart depicting the Information Gain and Expected Reward Gain calculated for $A2$ in two cases: b=$U(0,1)$ (Blue) and b=$U(\frac{5}{12},1)$ (Orange) (b) Values of $J$ associated with actions $[A1,A2]$ from reward matrix given in Figure \ref{fig:information_sufficiency_example_reward_matrix} based on the given definition of $b$.}
\label{exp1_results}
\end{figure}

Consider the reward matrix presented in Figure \ref{fig:information_sufficiency_example_reward_matrix}. It can be calculated that the intersection point for $A1$ is $\alpha = \frac{5}{12}$, and for $A2$ is $\alpha=\frac{5}{6}$ (for clarity, in both cases this means that if $C$ chooses $B1$, this indicates $\alpha$ is greater than the intersection value). The values for $J$ based on the Information Gain and Expected Reward Gain definitions are given in Table \ref{table:exp1_results}. 

We first consider when $b^{0} = U(0,1)$. In this case, action $A1$ has the potential to determine that $\alpha \in [\frac{5}{6},1]$, which would be a greater gain in information than what could result from choosing $A2$, $\alpha \in [0,\frac{5}{12}]$. This is reflected in $A1$ and $A2$ having associated Information Gain values of $0.68$ and $0.45$ respectively. In terms of expected reward gain, similarly, since $A1$ has a high likelihood of receiving the extreme positive reward ($5$), while the most likely outcome from $A2$ being a reward of $0$, $A1$ and $A2$  have expected reward gain values of $3.54$ and $1.25$ respectively. 

We next repeat this calculation again, but with $b = U(\frac{5}{12},1)$ (this would arise if $R$ had chosen $A1$ and observed $C$ perform $B1$). After this observation, there is no informational benefit to choosing $A1$, since it is already known that $\alpha$ is within the range associated with the action. In terms of expected reward gain, there is also little reason to choose $A2$, since $R$ already knows it can receive the optimal reward ($5$) by choosing $A1$. Therefore this action receives a significantly reduced value for $J$, as compared to the previous case ($0.41$ compared with $1.25$). Since the knowledge $R$ has is \emph{sufficient} for it to receive it's reward, $J$ no longer provides significant motivation to explore. On the other hand, $A2$ can still provide a lot of information about the value of $\alpha$, since it is possible to determine if $\alpha \in [\frac{5}{6},1]$, so its information gain value increases compared to the previous case ($0.6$ compared with $0.45$). Even though this extra information is of no value to $R$, the information gain value cannot capture this, and so $R$ is still incentivised to explore. Figure \ref{fig:exp1_results} presents a stacked bar chart reflecting this graphically. 

We refer to the diminishing of the motivation to explore as the resulting information ceases to be beneficial as \emph{Information Sufficiency}; any reward function that motivates exploration should incorporate this sufficiency in order to limit unnecessary exploration. Metrics such as Information Gain motivate pursuing certainty for certainty's sake, which does not always align with the decision-maker's motivation, to receive as high a reward as possible. Expected Reward Gain aligns with the decision-maker's motivations, as it only assigns high values when there is the possibility of achieving a higher reward. 

\section{Active Altruism Learning for Autonomous Driving}
In the previous section we demonstrated how Active and Online learning methods could be used to motivate an interactive agent to choose actions that would reveal information about another agent's altruistic tendencies, thereby allowing it to pursue a preferred equilibrium. This observation required the assumption that a player's actions could be observed perfectly, and that actions were executed instantaneously. In this section we will demonstrate how, by relaxing these assumptions, the proposed decision-making system can be used in autonomous driving settings to motivate the execution of information gathering trajectories.

\subsection{Experimental Setup}
The setting for this experiment is the lane change setting introduced in Figure \ref{intro_figure}. For demonstrative purposes we use an alternative game matrix to motivate decision-making (Figure \ref{active_alt_exp2_figure_table}). 

\begin{figure}[!h]
    \begin{center}
        \begin{tabular}{cc|c|c|}
            & \multicolumn{1}{c}{} & \multicolumn{2}{c}{$C$}\\
            & \multicolumn{1}{c}{} & \multicolumn{1}{c}{Behind}  & \multicolumn{1}{c}{Ahead} \\\cline{3-4}
            \multirow{2}*{$R$}  & $A$ & $(3,-2)$ & $(-10,3)$ \\\cline{3-4}
            & $B$ & $(0,-2)$ & $(1,3)$ \\\cline{3-4}
            & $E$ & $(2,0)$ & $(-1,3)$ \\\cline{3-4}
        \end{tabular}
    \end{center}
\caption{Reward matrix associated with the Conflict-free lane change scenario.}
\label{active_alt_exp2_figure_table}
\end{figure}

At the end of the interaction both agents would prefer to be ahead of the other agent. $R$, achieves this outcome only if they attempt to merge ahead ($A$), and $C$ facilitates it. If the $C$ chooses to remain ahead (Ahead) in their lane, then they will stay ahead regardless of what $R$ does. Thus they always receive the optimal reward for choosing to stay ahead. 

While the optimal reward for $R$ occurs if they attempt to merge ahead, this is also associated with the highest penalty as, if they attempt to merge ahead and $C$ does not facilitate it, ($A$,Ahead), then the player executing the manoeuvre is being reckless, and is punished accordingly. $R$'s merge behind manoeuvre ($B$) is ``safe", since $C$ will always choose to stay ahead in this case (($B$,Ahead) strictly dominates ($B$,Behind)). Therefore this is an uninformative action. Finally, $R$'s ``exploratory action" ($E$) is designed to return information about the $C$'s preferences, without actually pursuing the objective. An example of such an action would be turning on an indicator and observing how $C$ responds, or nudging into the lane. This returns a positive reward if $C$'s response indicates that the $R$ can merge ahead, and a negative reward if the response indicates otherwise.

As in the previous section $\alpha_{R}=0$ and is known by both players. $C$ has an unknown altruism coefficient, $\alpha$. $R$, instigating the manoeuvre, is also known to be the leader in the game. 

As in \cite{geary2021}, planning is performed by trajectory optimisation; each cell, ($Ai$,$Bj$), of the game matrix corresponds to a pair of weight vectors, ($\vec{w}_{ijR}$,$\vec{w}_{ijC}$), and the cost function player $k$ optimises over, $J_{k}$, is defined as; $J_{k} = w_{ijk}\vec{\phi}^{T}(\vec{x_{k}},\vec{x_{-k}})$, $w_{ijk} \in \mathbb{R}^{K}$, where $K=6$ is the number of cost function features. In this experiment the features, $\vec{\phi}$, are defined as:
\begin{itemize}
    \item $\phi_{0} = 1-\exp(\lambda_{x}(x_{k}-x_{l})^{2})$; squared lateral distance from left lane centre.
    \item $\phi_{1} = 1-\exp(\lambda_{x}(x_{k}-x_{r})^{2})$; squared lateral distance from right lane centre.
    \item $\phi_{2} = 1-\exp((v_{k}-v_{\text{limit}})^{2})$; squared difference between velocity and speed limit.
    \item $\phi_{3} = 1-\exp(\lambda_{\theta}(\theta_{k} - \theta_{\text{lane}})^{2})$; squared difference between heading and lane heading.
    \item $\phi_{4} = -(1-(\frac{(x_{k}-x_{-k})\cos(\theta_{-k}) + (y_{k}-y_{-k})\sin(\theta_{-k})}{W+\epsilon})^{2}+$ \\* $(\frac{(x_{k}-x_{-k})\sin(\theta_{-k}) + (y_{k}-y_{-k})\cos(\theta_{-k})}{L+\delta})^{2})$; safety ellipse with major and minor axes $L+\delta$ and $W+\epsilon$, centred on the position of the $-k$ vehicle.
    \item $\phi_{5} = \tanh(y_{k}-y_{-k})$; agent $k$ receives a positive reward for being ahead of agent $-k$, and a negative reward for being behind.
\end{itemize}

where agent $k$'s state $\vec{x}_{k} = [x_{k},y_{k},v_{k},\theta_{k}]$ are the (x,y)-coordinates, linear velocity and heading respectively. $W$ and $L$ are the vehicles' width and length. $\epsilon$,$\delta$, $\lambda_{x}$ and $\lambda_{\theta}$ are parameters whose values were empirically chosen. Each lane is of width $5$ metres.

Following the approach used in \cite{geary2021}, we perform Model Predictive Control (MPC) to generate the trajectories. However, unlike in those experiments, we use the trajectory generation method proposed in \cite{sadigh2016,sadigh2018}, which generates the trajectory using bi-level optimisation; 

\begin{equation}
    \begin{split}
        u^{*}_{R},u^{*}_{C} &= \argmax_{u_{R}} J_{R}(x_{R},x_{C},u_{R},u_{C}^{*}(x_{C},x_{R},u_{R}))\\
        & s.t. \text{ } u_{C}^{*}(x_{R},x_{C},u_{R}) = \argmax_{u_{C}} J_{C}(x_{C},x_{R},u_{C},u_{R})
    \end{split}
\end{equation}

$u^{*}_{k} = \{\vec{u}^{t}_{k}\}_{t=0}^{T}, k\in \{R,C\}$ and the Kinematic Bicycle dynamics model, \cite{kong2015}, is applied to define $\{x^{t}_{k}\}_{t=0}^{T}$. This model treats $R$ as the leader in planning as well as in decision-making, as it is assumed that the $C$ generates their trajectory optimally in response to $R$'s actions. MPC is performed with timestep $dt=.2$ seconds, and a lookahead horizon of $T=1.2$ seconds. Trajectories are recomputed every timestep, and the experiment ends after $30$ timesteps. The weight values $\vec{w}_{ijk}$ were set empirically based on the behaviour specified by the corresponding action combination.

Unlike in the previous section, where $R$ had perfect observability of $C$'s actions, in this experiment $R$ does not know the true action, $B_{C}$, chosen by $C$. $C$, as the follower, can observe $R$'s action and responds optimally. To account for this uncertainty we use Bayes' Rule to update, $b(\alpha)$:

\begin{equation}
    \begin{split}
        b^{t+1}&(\alpha \in D) = P(\alpha \in D|Ai,u_{C}^{0:t})\\
        &\propto \sum_{B_{C} \in B} P(B|Ai,u_{C}^{t})P(B|Ai,\alpha \in D)b^{t}(\alpha \in D)\\
        &= P(B_{C}^{D}|Ai,u_{C}^{t})b^{t}(\alpha \in D)
    \end{split}
\end{equation}

$\forall D = [c,d] \in \Delta$, $c,d \in \mathbb{R}$, where $\Delta$ is defined as the set of all disjoint ranges which result from partitioning $[0,1]$ using the intersection of reward function values (as detailed in Section \ref{parameter_inference_for_stackelberg_games}). $Ai$ is $R$'s chosen action, and $u_{C}^{0:t}$ are from $C$'s observed trajectory. $B_{C}^{D}$ is the optimal game action for $C$ to choose, given $\alpha \in D$. The third line follows from the second as, if it is known that $\alpha \in D$, $P(B_{C}^{D}|\alpha \in D) = 1$. The only unknown quantity in the equation is $P(B_{C}^{D}|Ai,u_{C}^{t})$, which we define using the softmax function:

\begin{equation}
    P(B_{C}^{D}|Ai,u_{C}^{t}) = \frac{\exp(\vec{w}_{ijC}\vec{\phi}(u_{C}^{t}))}{\sum_{k}\exp(\vec{w}_{ikC}\vec{\phi}(u_{C}^{t}))}
\end{equation}

where $j$ is the index of $B_{C}^{D}$ in $B$. As in the previous section we start each experiment with a uniform belief over the value of $\alpha$; $b^{0}(\alpha) = U(0,1)$. In the next section we present the results when this approach is applied to the lane merge scenario.

\subsection{Experiment Results}
From Figure \ref{active_alt_exp2_figure_table} we can determine how the reward matrix partitions the range of possible values for $\alpha$; we note that $A$ has an intersection value of $\frac{5}{18}$, $B$ $\frac{5}{4}$ and $E$ $\frac{1}{2}$. From this we can conclude that $\Delta = \{[0,\frac{5}{18}],[\frac{5}{18},\frac{1}{2}],[\frac{1}{2},1]\}$. $\alpha \in [0,\frac{5}{18}]$ indicates that the ($A$,Behind) equilibrium is infeasible, so $R$ should choose $B$. Otherwise $R$ should choose $A$. In this section we will run the experiment detailed previously for $\alpha \in [0,\frac{5}{18}]$ ($\alpha = 0.2$) and for $\alpha \in [\frac{5}{18},1]$ ($\alpha = 0.9$)\footnote{The magnitude of $\alpha$ within the domain does not affect the outcome. Values are chosen for clarity.}. In Sections \ref{active_altruism_learning_for_stackelberg_games} and \ref{information_sufficiency} we compared the performance of $3$ different definitions of the information gathering reward, $J$. In this section we will compare and evaluate these definitions by observing the resulting behaviour in each of the experiment cases.

\subsubsection{$\alpha \in [0,\frac{5}{18}]$:}
\begin{figure}[h]
    \centering
    \subfloat[\label{fig:active_learning_exp2_a_2_executed_trajectory}]{%
        \includegraphics[height=.2\textheight, width=.5\textwidth]{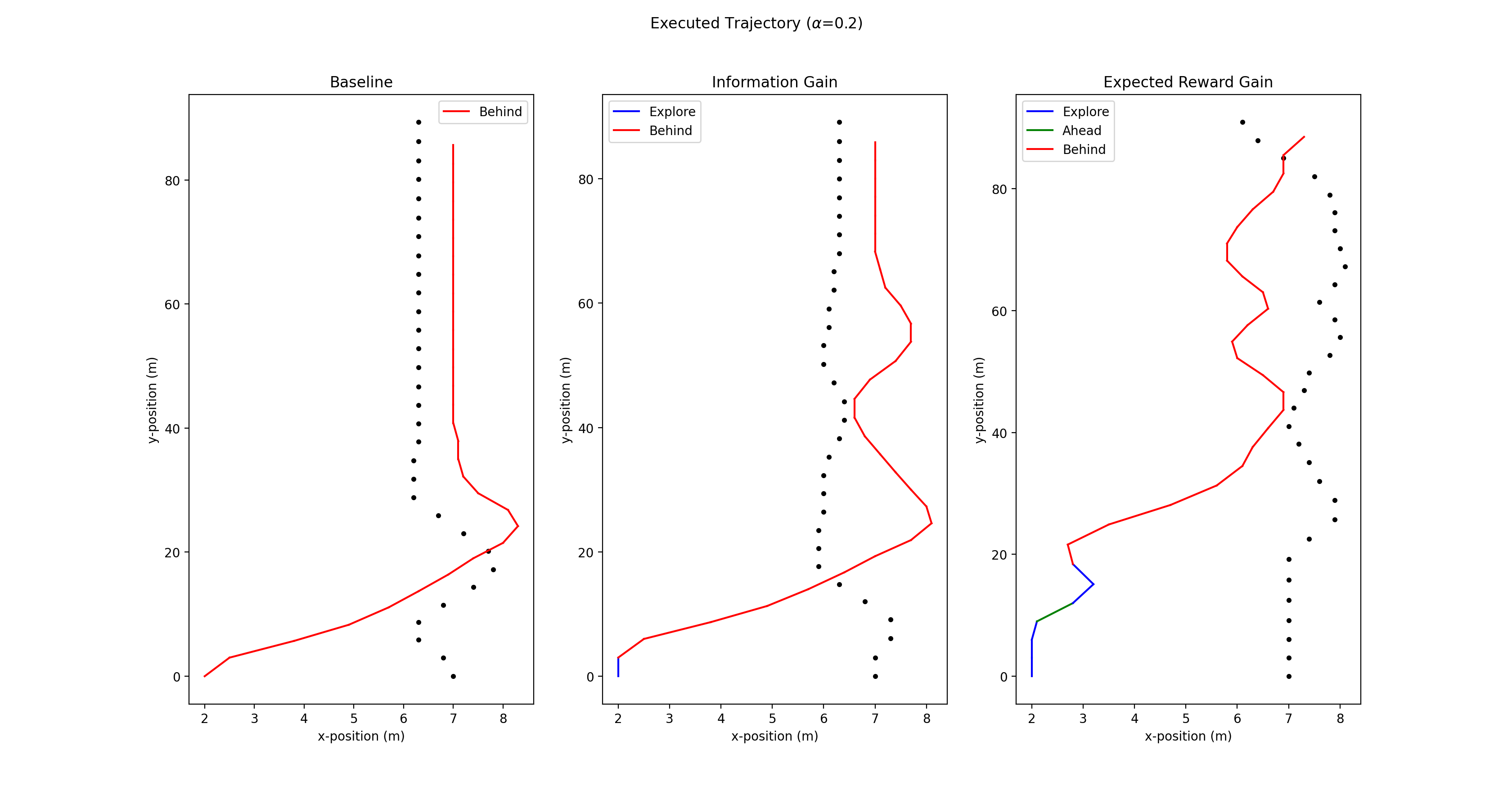}}
    \newline
    \subfloat[\label{fig:active_learning_exp2_a_2_relative_position}]{%
        \includegraphics[height=.2\textheight, width=.5\textwidth]{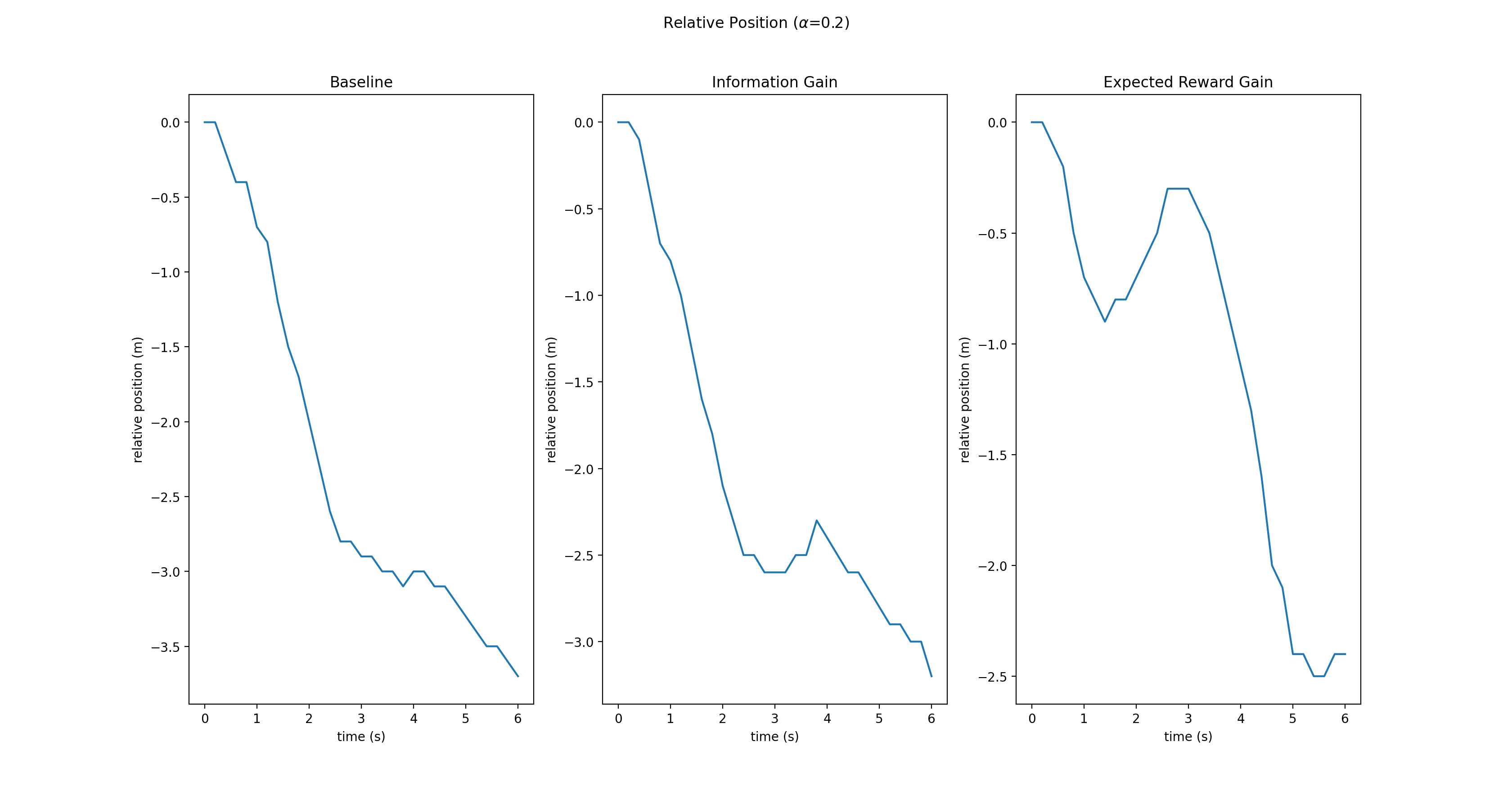}}
    \caption{Results for the information gathering trajectory generation experiment for $\alpha=0.2$; (a) Vehicle trajectories; $C$ is given by the dotted line. $R$ is given by the unbroken line. (b) Relative longitudinal position of $R$ with respect to $C$. (Left) Baseline ($J=0$), (Middle) Information Gain ($J=H(b^{t})-H(b^{t+1})$, (Right) Expected Reward Gain ($J=|F(b^{t+1})-F(b^{t})|$).}
\label{fig:active_learning_exp2_a_2_plot}
\end{figure}

At the lower end of the spectrum of possible altruism values, $C$ is considered to be relatively selfish, and will, per the reward matrix in Figure \ref{active_alt_exp2_figure_table}, never choose to allow the $R$ to merge ahead. As a result, if $R$ accurately determines the true value of $\alpha$, they will choose to merge behind. The results of this experiment are presented in Figure \ref{fig:active_learning_exp2_a_2_plot}. 

From Figure \ref{fig:active_learning_exp2_a_2_plot}(Left) we observe that, in the baseline case where $R$ is only motivated by the immediate expected reward of the action, the agent immediately chooses the ``safe" action of merging behind. This provides no information about the value of $\alpha$. As we will see in the next section, the fact that this baseline ultimately chooses the preferable action is purely incidental. 

Figure \ref{fig:active_learning_exp2_a_2_plot}(Middle) demonstrates the behaviour when $R$ is motivated by information gain. While information gain can motivate exploration of informative actions, it is unclear how to determine the value of the scaling parameter, $\lambda$, other than in an ad-hoc, case-specific basis. In these experiments we let $\lambda = 1$. From the results we observe that this was sufficient to motivate choosing the exploratory action, during which time $R$ observes whether $C$ is going to move forward, or give way. This provides information about $\alpha <0.5$. Once there is sufficient evidence of this fact, the agent chooses to merge behind. This does not explore the possibility that $\alpha \in [\frac{5}{18},\frac{1}{2}]$, even though this would allow for it to merge ahead. Therefore, while this method does some exploration, it does not explore all of the possible options. 

The final set of results, in Figure \ref{fig:active_learning_exp2_a_2_plot}(Right), depict the case when $R$ is motivated by expected reward gain. We observe that the agent explores initially, as in the previous case. Once the agent is sufficiently convinced that $\alpha<.5$, unlike the information gain case, it performs the $A$ action, since this is the only way to determine if $\alpha \in [\frac{5}{18},\frac{1}{2}]$, which would achieve a higher reward. Once it has been determined that this is not the case, the agent returns to the exploratory action, before converging on the merge behind manoeuvre. 

With this result we have demonstrated that, while terms such as information gain can motivate exploratory behaviour to a certain degree, they do not reliably explore all the possible outcomes. Of the tested models, our proposed expected reward gain motivation was the only one to choose actions so as to completely eliminate the possibility of achieving the optimal reward outcome. Next we will demonstrate how a thorough exploration of the possibilities can aid in achieving an optimal outcome.

\subsubsection{$\alpha \in [\frac{5}{18},1]$:}
\begin{figure}[h]
    \centering
    \subfloat[\label{fig:active_learning_exp2_a_9_executed_trajectory}]{%
        \includegraphics[height=.2\textheight,width=.5\textwidth]{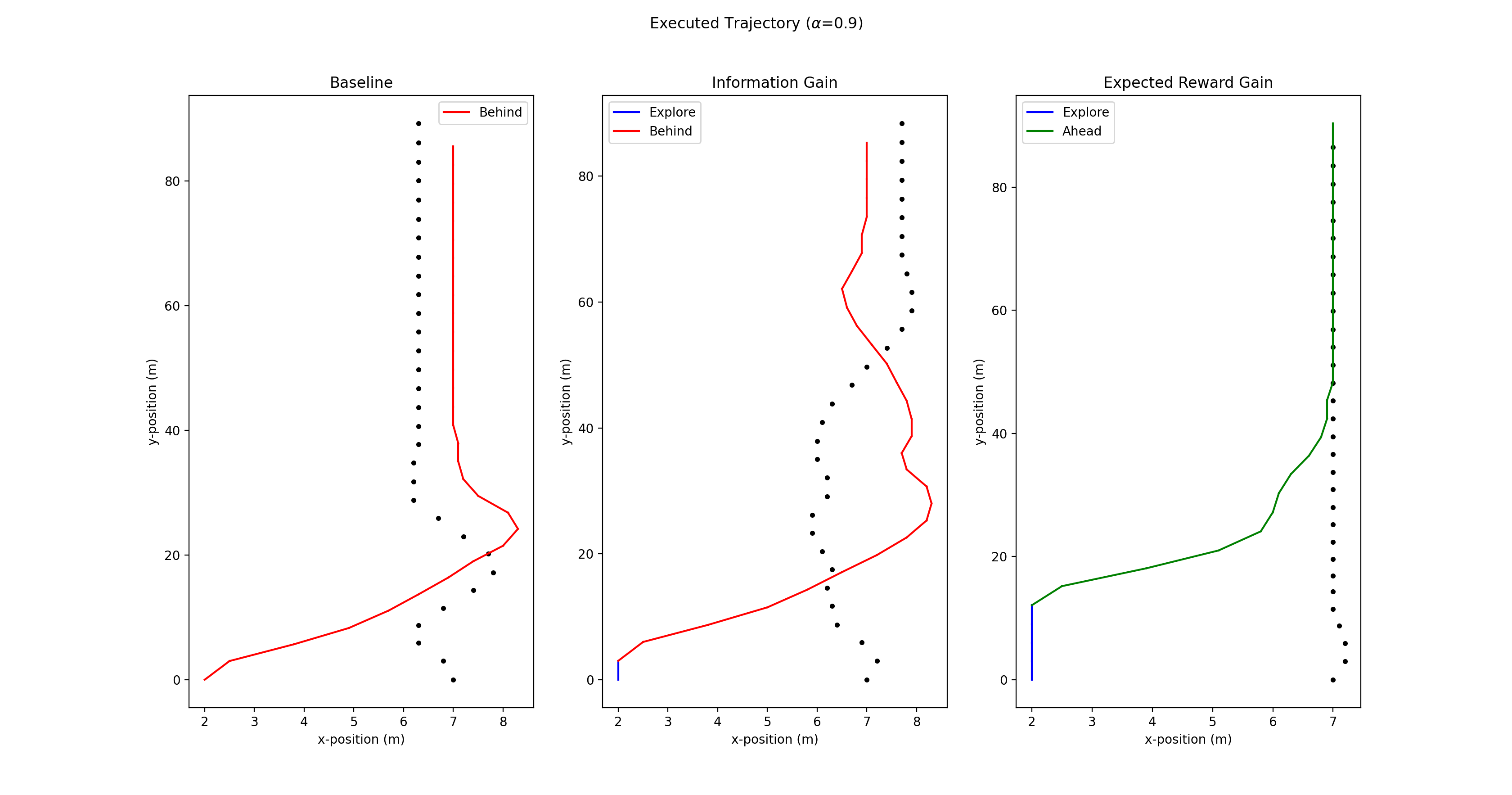}}
    \newline
    \subfloat[\label{fig:active_learning_exp2_a_9_relative_position}]{%
        \includegraphics[height=.2\textheight,width=.5\textwidth]{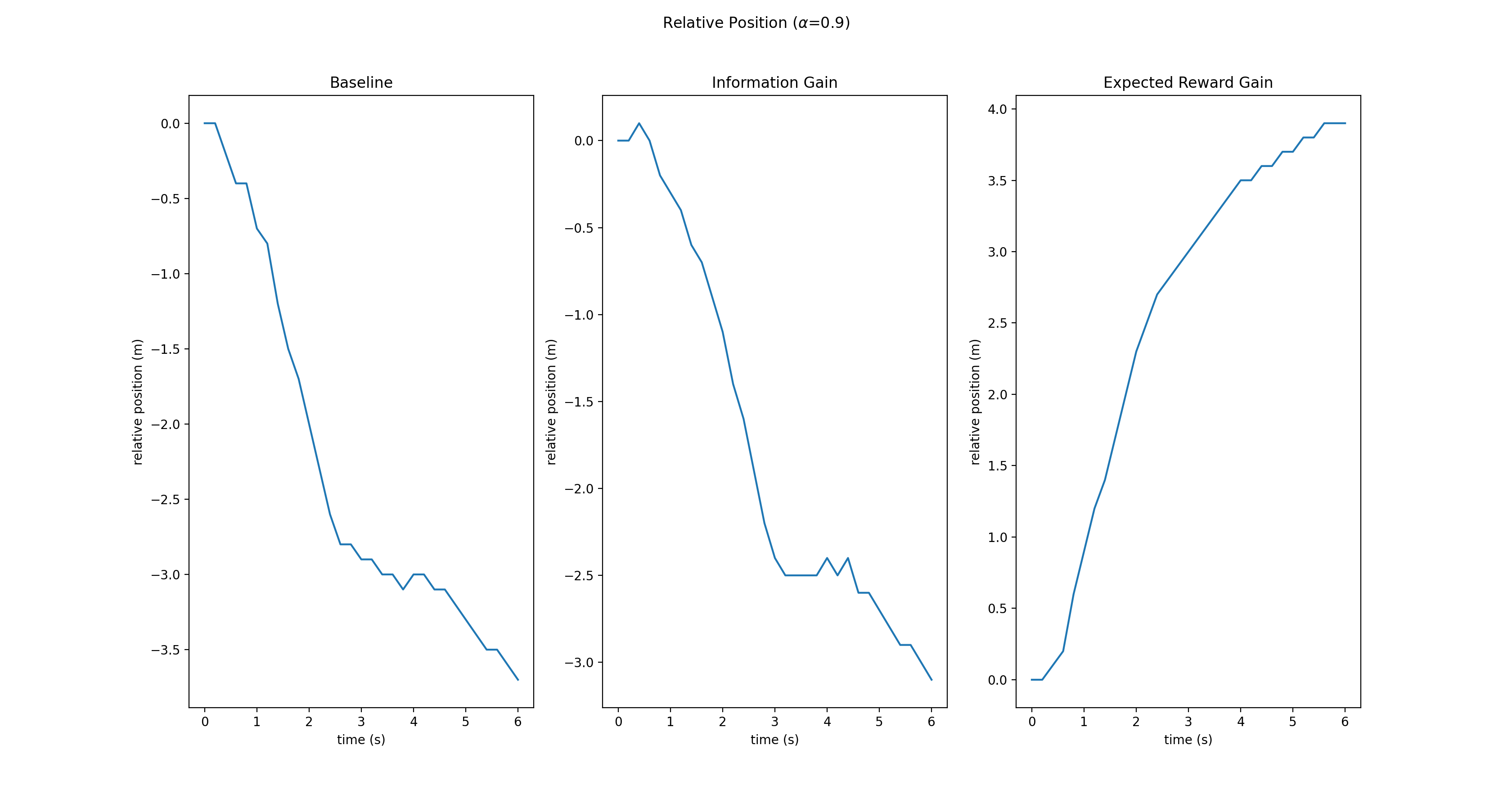}}
    \caption{Results for the information gathering trajectory generation experiment for $\alpha=0.9$; (Left) Vehicle trajectories; $C$ = dotted line. $R$ = unbroken line. (Middle) Relative longitudinal position $R$ with respect to $C$; (Right) Belief in value of $\alpha$. (a) Baseline ($J=0$), (b) Information Gain ($J=H(b^{t})-H(b^{t+1})$, (c) Expected Reward Gain ($J=|F(b^{t+1})-F(b^{t})|$).}
\label{fig:active_learning_exp2_a_9_plot}
\end{figure}

Since $\alpha > \frac{5}{18}$, we observe from Figure \ref{active_alt_exp2_figure_table} that, if $\alpha$ was known, $R$ could achieve their optimal reward by choosing $A$ and merging ahead. However, as we see in Figure \ref{fig:active_learning_exp2_a_9_plot}, this outcome is only realised when expected reward gain is used. The baseline case is as in the previous experiment, with $R$ immediately defaulting to the ``safe" course of action and merging behind.

In the case when information gain is used to motivate the exploration, $R$ initially chooses the exploratory action ($E$), which provides some information about the value of $\alpha$; we observe from Figure \ref{fig:active_learning_exp2_a_9_relative_position} (Middle) that $C$ does give way as a result to the exploration action (we observe that, on the plot, $R$ is briefly ahead of $C$, indicating that $C$ slowed down to give way). Paradoxically, however, as a result of this observation, $R$ chooses to merge \emph{behind} $C$. This can be explained as follows; when $b^{0}(\alpha) = U(0,1)$ initially, $\mathbb{E}[R_{R}^{*}(A)] = -.02$, $\mathbb{E}[R_{R}^{*}(B)] = 1$, and $\mathbb{E}[R_{R}^{*}(E)] = 1.19$. Therefore the explore action is only slightly more preferable than simply merging behind, while the riskier action to attempt and merge ahead is significantly less desirable. After performing the exploration action, and gaining information about $\alpha$, the value of $J(E)$ goes down. But this is not commensurate with the increase in $\mathbb{E}[R_{R}(E)]$, so the overall reward, $R_{R}^{*}$, falls beneath the value of $1$, and so merging behind becomes preferable, at which point no further information is gained. Since $R$ never gains enough confidence that $\alpha>\frac{5}{18}$ to overcome the risk of being wrong, $A$ is never considered as an action. 

In the case when $R$ is motivated by expected reward gain, $R$ spends longer performing the exploration action. Once it has gained sufficient certainty that $\alpha>\frac{5}{18}$ it determines that it can safely merge ahead of $C$'s vehicle, achieving the optimal equilibrium. 
\section{Conflict-aware Active Altruism Learning}
In the context of autonomous driving, the bimatrix component of the hierarchical model approximates the \emph{common} awareness that drivers have about a driving situation. Therefore, each driver must be able to independently construct equivalent game matrices. The previous experiments utilised two assumptions that contradict this requirement; hand crafted game matrix values were used, and the planning agent was presumed to be the leader in the Stackelberg Game. In the open world, if the vehicles are unable to communicate directly, there would be no way to satisfy these requirements. In this section we will propose a method for relaxing these assumptions, so that active-altruism learning can be effectively implemented without relying on known roles or hand-crafted matrix values.

In previous sections the magnitudes of the game matrix values were defined based on the outcome expected by the context, and the behaviour being explored in the experiment. In general, a standard method for defining these values must be established in order to ensure that vehicles can produce equivalent game matrices. \cite{Shalev-Schwartz2017} proposes a comprehensive set of criteria for assigning culpability ("responsiblity") in a driving accident resulting from behavioural error. This includes metrics for evaluating responses in merging scenarios. Based on these criteria we propose a trinary reward function based on accident responsibility:
\begin{equation}
  r_{ijX} =
  \begin{cases}
    -1 &\text{ if $(Ai,Bj)$ is an accident and X is responsible} \\
    1 &\text{ if $(Ai,Bj)$ achieves X's goal}\\
    0 &\text{ otherwise}
  \end{cases}  
\end{equation}
for $X \in \{R,C\}$. This reward captures the common awareness drivers have about the motivations of other drivers. The degree to which each driver wants a particular outcome to occur is captured by their $\alpha$ coefficient. We use these rules to construct Figure \ref{intro_figure_table} for the lane merge scenario. For example, when ($A$,Behind) is executed, $R$ ends up ahead, receiving a reward of $1$, whereas $C$ is no worse off than before, so gets a reward of $0$. In the case of ($A$,Ahead), both cars would be to blame for the resulting accident, so both receive $-1$. Since these rules do not depend on any communication between vehicles, they can be computed individually by each player.

In our experiments it was assumed that the Leader Follower roles had been established. In practice, without a means of communication, this would not be the case (e.g., if both cars were controlled by our proposed model, both would presume to be the leader). This can result in a breakdown in the ability to coordinate that \cite{geary2021} call ``Conflict". For example, in Figure \ref{intro_figure_table}, if $C$ believes they are the leader they will choose $A$, expecting $C$ to choose Behind. But, if $C$ believes they are the Leader they would choose Ahead. Therefore, Conflict should be accounted for in such models.

We can account for Conflict by considering regions of the domain of $\alpha$ where Conflict would result in a coordination breakdown; for $\alpha_{R}=0$, this occurs for $\alpha_{C}<0.5$. In this region $R$ should be prepared for the possibility that $C$ will behave as if they were the leader, and choose their preferred action. To account for this awareness we define;
\begin{equation}
    \begin{split}
        \tilde{R}_{R}(Ai,Bj,\alpha) = (1-&~b^{t}(\alpha<0.5))R_{R}(Ai,Bj,\alpha)+ \\
        &b^{t}(\alpha<0.5)R_{R}(Ai,Bj^{'},\alpha)
    \end{split}
    \label{conflict_aware_reward}
\end{equation}
where $b^{t}(\alpha<x)$ is the probability $\alpha<x$ based on $R$'s belief at time $t$, and $Bj^{'}$ is the action $C$ would choose if they were leader. 
\begin{figure}[t]
    \centering
    \includegraphics[height=.2\textheight,width=.5\textwidth]{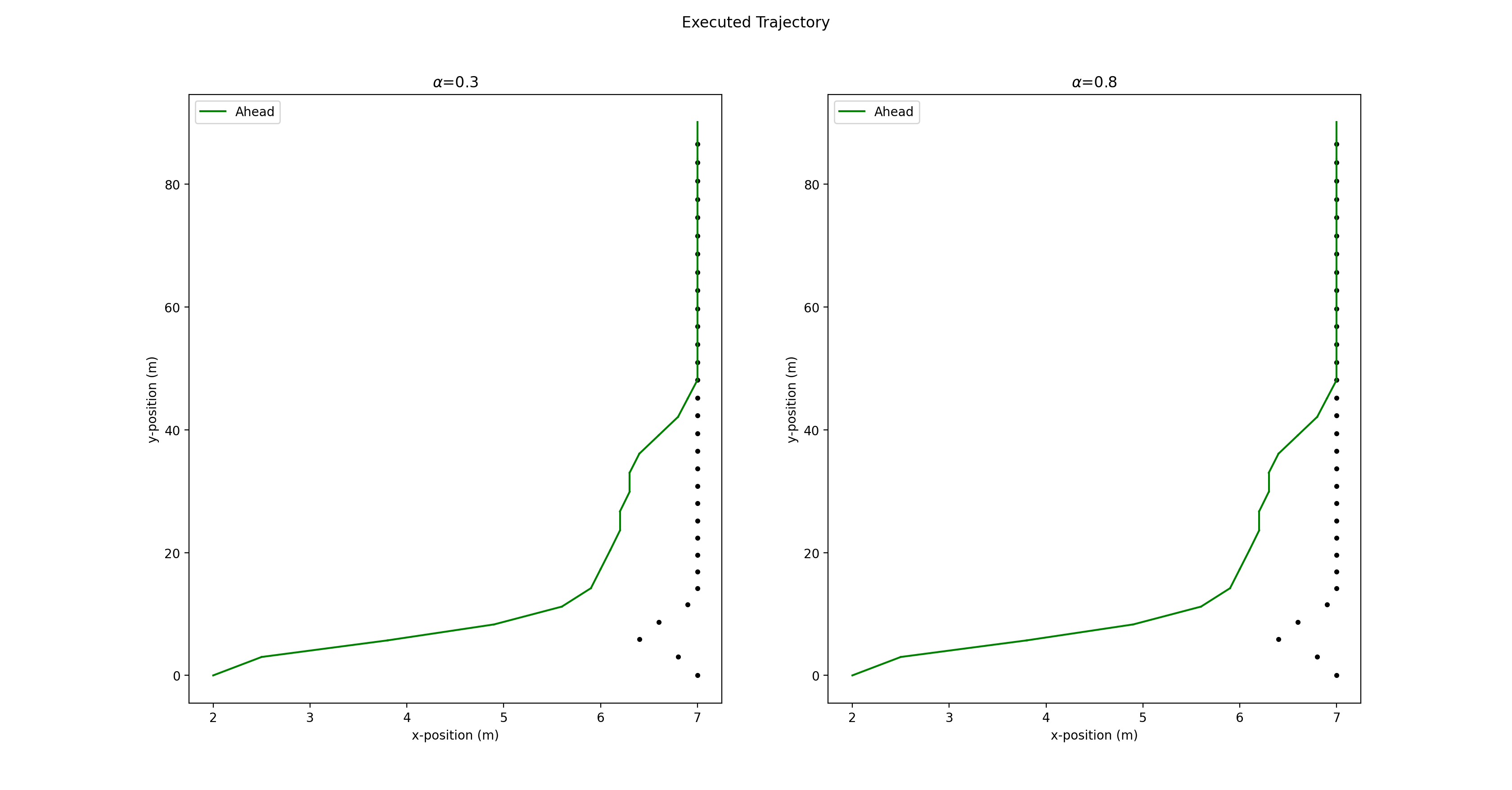}
    \caption{Conflict-unaware trajectory execution; $C$ = dotted line. $R$ = unbroken line.}
    \label{fig:conflcit_unaware_plot}
\end{figure}

Figure \ref{fig:conflcit_unaware_plot} depicts the trajectories produced from the standard definition of $R_{R}$, using the matrix values in Figure \ref{intro_figure_table}, and the expected reward gain motivation to perform decision-making without accounting for Conflict; $R$ always chooses $A$, expecting $C$ to oblige.\footnote{For technical reasons in these experiments $C$'s executed trajectory presumes $R$ is the leader. Their behaviour is not relevant to our discussion.} If $C$ had assumed they were the leader, then they would have chosen Ahead, and each vehicle would believe the other was behaving sub-optimally. This belief could result in an accident, or a stalemate as each vehicle believes the other is in the wrong. Without communication between the vehicles, there is no clear way to resolve this outcome, so it is preferable that it be avoided entirely.
\begin{figure}[t]
    \centering
    \includegraphics[height=.2\textheight,width=.5\textwidth]{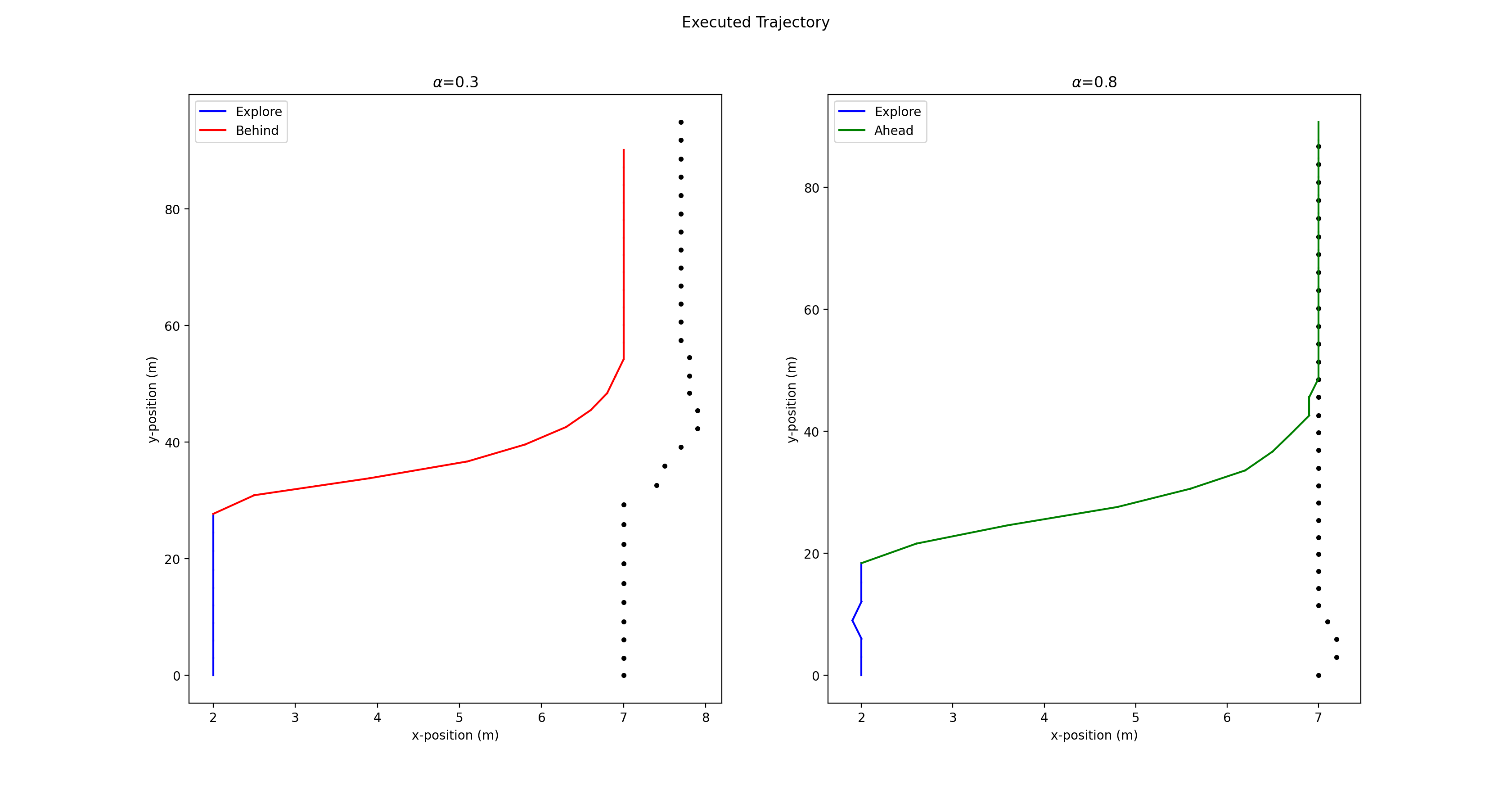}
    \caption{Conflict-aware trajectory execution; $C$ = dotted line. $R$ = unbroken line.}
    \label{fig:conflcit_aware_plot}
\end{figure}

Figure \ref{fig:conflcit_aware_plot} presents the trajectories produced under the same settings when Conflict is accounted for by using $\tilde{R}_{R}$ (Equation \ref{conflict_aware_reward}) in decision-making in the place of $R_{R}$. With this awareness, in both cases $R$ initially chooses $E$, to test for the value of $\alpha$. In the conflicted case, Figure \ref{fig:conflcit_aware_plot}(Left), $R$ gives way, due to the risk of an accident due to Conflict. In the conflict-free case $R$ knows that Conflict does not change the outcome, and chooses $A$, knowing that their intent will not be misinterpreted and will be facilitated.

These results demonstrate that, using our proposed standard for defining game matrices, and our method for accounting for conflict, an AV is able to effectively utilise active learning based methods to reliably and efficiently infer the value of $\alpha$ to a sufficient degree to be able to navigate the lane merge scenario. 
\section{Conclusion}
In this work we have proposed a method for incorporating Active Learning methods into a hierarchical trajectory generation model for autonomous driving applications. We defined ``Information Sufficiency", and demonstrated that reward functions that do not account for this are prone to inadequate exploration and sub-optimal behaviour. We proposed a novel reward function, Expected Reward Gain, that was conscious of Information Sufficiency. We showed that, in part due to this awareness, the reward function was better suited to motivate useful exploration. We proposed a standard for constructing game matrices based on accident culpability, as well as a method to account for uncertainty in the roles of leader and follower in the resulting game. Finally we demonstrated how the proposed standard could be applied to the lane merge problem to motivate appropriate and effective inference of the $\alpha$ parameter value.



\bibliographystyle{ACM-Reference-Format} 
\bibliography{references.bib}


\end{document}